\documentclass[10pt,twocolumn,letterpaper]{article}
\usepackage[pagenumbers]{cvpr}
\usepackage[dvipsnames]{xcolor}

\definecolor{cvprblue}{rgb}{0.21,0.49,0.74}
\usepackage[pagebackref,breaklinks,colorlinks,citecolor=cvprblue]{hyperref}
\usepackage{pgfplots}
\usetikzlibrary{calc}
\pgfplotsset{compat=1.18}
\usepackage[numbers]{natbib}
\usepackage{arydshln}
\usepackage{wrapfig,lipsum,booktabs}
\usepackage{multirow}
\usepackage{color}
\usepackage{multicol}
\usepackage{algorithm}
\usepackage{listings}
\usepackage{xcolor}
\usepackage{graphicx}
\newcommand{\knn}{$k$-NN\xspace}
\newcommand{\norm}[1]{\left\lVert#1\right\rVert}

\title{Guarding Barlow Twins Against Overfitting with Mixed Samples}

\author{Wele Gedara Chaminda Bandara$^1$, \hspace{3pt} Celso M. De Melo$^2$, \hspace{3pt} and Vishal M. Patel$^1$\\
$^1$Johns Hopkins University\hspace{16pt}$^2$US Army DEVCOM Research Laboratory\\
{\tt\small \url{https://github.com/wgcban/mix-bt.git}}}

\begin{document}
\maketitle
\begin{abstract}
Self-supervised Learning (SSL) aims to learn transferable feature representations for downstream applications without relying on labeled data. The Barlow Twins algorithm, renowned for its widespread adoption and straightforward implementation compared to its counterparts like contrastive learning methods, minimizes feature redundancy while maximizing invariance to common corruptions. Optimizing for the above objective forces the network to learn useful representations, while avoiding noisy or constant features, resulting in improved downstream task performance with limited adaptation.
Despite Barlow Twins' proven effectiveness in pre-training, the underlying SSL objective can inadvertently cause feature overfitting due to the lack of strong interaction between the samples unlike the contrastive learning approaches. From our experiments, we observe that optimizing for the Barlow Twins objective doesn't necessarily guarantee sustained improvements in representation quality beyond a certain pre-training phase, and can potentially degrade downstream performance on some datasets.
To address this challenge, we introduce Mixed Barlow Twins, which aims to improve sample interaction during Barlow Twins training via linearly interpolated samples. This results in an additional regularization term to the original Barlow Twins objective, assuming linear interpolation in the input space translates to linearly interpolated features in the feature space. Pre-training with this regularization effectively mitigates feature overfitting and further enhances the downstream performance on CIFAR-10, CIFAR-100, TinyImageNet, STL-10, and ImageNet datasets. The code and checkpoints are available at: \url{https://github.com/wgcban/mix-bt.git}
\end{abstract}    
\section{Introduction}
\label{sec:intro}
\begin{figure}[tbh!]
    \centering
    \includegraphics[width=\linewidth]{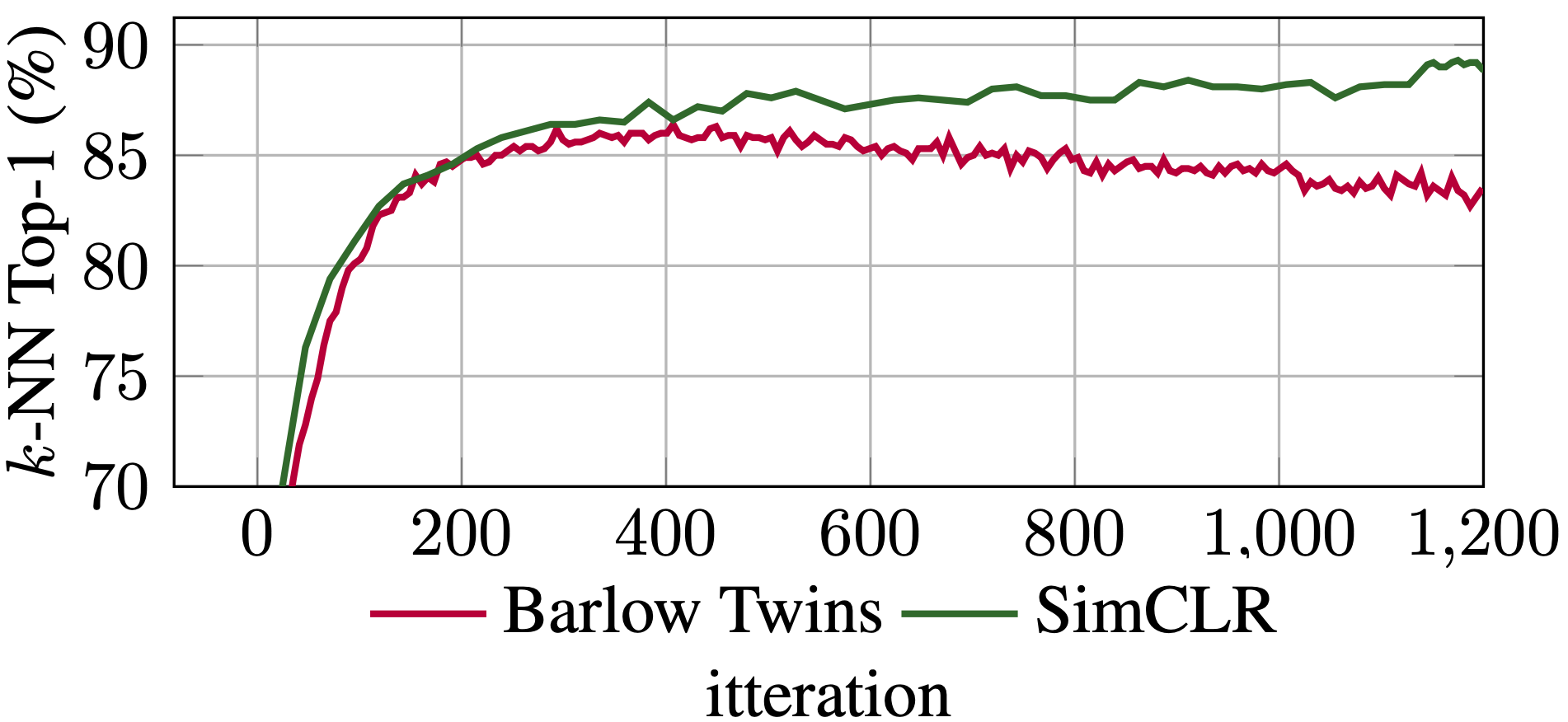}
    \caption{Assessing the representation quality via \knn accuracy on the test-set during SSL training on train-set for information maximization-based \textcolor{purple}{\bf Barlow Twins}~\cite{zbontar2021barlow} \textit{vs.} contrastive learning-based \textcolor{OliveGreen}{\bf SimCLR}~\cite{chen2020simple}, on CIFAR-10~\cite{krizhevsky2009learning} dataset.}
    \label{fig:intro-overfit}
\end{figure}

Self-Supervised Learning (SSL) has experienced remarkable advancements in recent years~\cite{bachman2019learning, misra2020self, he2020momentum, tian2020makes, caron2020unsupervised, grill2020bootstrap, chen2021exploring, colorful, Doersch_2015_ICCV}, consistently outperforming supervised learning across numerous downstream tasks~\cite{zhong2022self}. Among the various SSL techniques, joint embedding architectures have garnered significant attention~\cite{assran2023self}. In these architectures, two networks are trained to generate similar embeddings for different perspectives of the same image. A prominent example is the Siamese network architecture~\cite{bromley1993signature, chen2021exploring}. However, Siamese network architectures are susceptible to \textit{``representation collapse.''} This occurs when the network disregards its inputs, leading to the generation of identical or irrelevant feature representations \cite{liu2021understand, liu2021self}. To address this issue, prior research has adopted two approaches: \textit{contrastive learning} and \textit{information maximization}.

Contrastive-learning~\cite{bromley1993signature, chopra2005learning, he2020momentum, hjelm2018learning, chen2020simple, tian_what, wu2018unsupervised, chuang2022robust} often entails substantial computational demands, necessitating large batch sizes~\cite{chen2020simple} or the utilization of memory banks~\cite{he2020momentum}. These methods employ a loss function designed to explicitly encourage the convergence of embeddings for similar images (positive samples) while pushing apart embeddings for dissimilar images (negative samples). 
 However, recent trends in SSL have shifted towards non-contrastive methods due to their simplicity of implementation and their ability to learn high-quality representations.  Notable examples of these non-contrastive SSL methods include BYOL~\cite{grill2020bootstrap} and SimSiam~\cite{chen2021exploring}. These methods employ various techniques such as batch-wise normalization, feature-wise normalization, ``momentum encoders'' 
~\cite{grill2020bootstrap, richemond2020byol}, and stop-gradient in one of the branches.

More recently, methods focused on preventing feature collapse through information maximization (InfoMax) have exhibited promising outcomes~\cite{zbontar2021barlow, ermolov2021whitening, bardes2021vicreg, bardes2022vicregl, wang2022rethinking}. These approaches aim to decorrelate every pair of variables within embedding vectors, thereby indirectly maximizing the information content of these vectors. The Barlow Twins~\cite{zbontar2021barlow} achieves this objective by aligning the cross-correlation matrix of the two embeddings with the identity matrix. Meanwhile, Whitening-MSE~\cite{ermolov2021whitening} whitens and spreads out the embedding vectors on the unit sphere. VICReg~\cite{bardes2021vicreg} and its variant VICRegL~\cite{bardes2022vicregl} introduce two regularization terms to maintain the variance of each embedding dimension above a threshold and decorrelate each pair of variables.

In this work, we revisit Barlow Twins~\cite{zbontar2021barlow}, one of the most widely adopted methods in SSL, and shed light on a critical aspect that demands attention. Our investigation reveals that the Barlow Twins, despite its notable strengths, is susceptible to \textit{overfitting}, particularly when the dimension of the embeddings experiences substantial growth. Our experiments conducted on small to medium-scale datasets, including CIFAR-10~\cite{krizhevsky2009learning}, CIFAR-100~\cite{krizhevsky2009learning}, TinyImageNet~\cite{le2015tiny}, and STL-10~\cite{coates2011analysis}, demonstrate that the \knn evaluation~\cite{devroye1996consistency} performance saturates or even deteriorates during the SSL training as the embedding dimensionality increases, as depicted in Figure \ref{fig:intro-overfit}. However, this is not the case with contrastive learning methods like SimCLR~\cite{chen2020simple}. This phenomenon indicates a tendency towards overfitting to the training data in Barlow Twins, with the model potentially memorizing specific instances and excessively focusing on feature representations~\cite{wang2022rethinking}. This, in turn, adversely affects the generalization of features for downstream applications. By evaluating the learned features with \knn, we can directly assess the quality of the learned representations without adapting them to a specific downstream task. These experiments conducted on datasets of varying sizes enable us to monitor the performance throughout the SSL process, leading to these critical insights.

Having identified the overfitting phenomenon in Barlow Twins, we explore various techniques for mitigating it. Notably, we discovered that MixUp regularization~\cite{zhang2017mixup}, a technique commonly employed in supervised learning, proves effective. This technique involves the linear mixing of two samples in the input space, and we formulate a regularization loss by establishing a relationship between the cross-correlation matrix of the mixed embeddings and the unmixed embeddings, under the assumption of the same linear interpolation in the embedding space. To the best of our knowledge, this marks the first utilization of MixUp regularization in InfoMax-based SSL, as existing SSL augmentations are typically applied on a per-sample basis. Our experiments conclusively demonstrate that the proposed MixUp-based regularization acts as a safeguard against overfitting in Barlow Twins. Furthermore, it leads to substantial improvements in performance on downstream applications, highlighting its capacity to facilitate the learning of high-quality features that significantly benefit a range of downstream tasks. In summary, this paper makes the. following contributions:
\begin{itemize}
    \item We revisit the Barlow Twins algorithm and identify its \textit{susceptibility to overfitting}.
    \item Through experiments on various datasets, we highlight the \textit{overfitting phenomenon when the embedding dimensionality increases}.
    \item We counteract feature overfitting in Barlow Twins with \textit{mixed sample interaction}, named Mixed Barlow Twins.
    \item Our experiments demonstrate that Mixed Barlow Twins \textit{improves the performance on downstream tasks over the Barlow Twins} and other state-of-the-art (SOTA) methods.
\end{itemize}
\section{Related Work}
\label{sec:related_work}
\paragraph{Contrastive Learning for SSL.} Contrastive learning methods, often employed in joint embedding architectures~\cite{Assran_2023_CVPR, Barchid_2023_CVPR, bardes2023mc}, aim to \textit{bring the output embeddings of two views of a sample closer to each other, while pushing other samples and their distortions farther apart.} This is typically achieved through the use of the InfoNCE loss~\cite{oord2018representation}. Such methods are commonly implemented using a Siamese network architecture, where the weights of the two branches are shared~\cite{misra2020self, he2020momentum, bromley1993signature, hjelm2018learning, chen2020simple, hadsell2006dimensionality, ye2019unsupervised, wu2018unsupervised, oord2018representation, chen2020big, chenempirical}. While contrastive learning techniques have demonstrated excellent performance, they come with a notable drawback - the requirement for a substantial number of contrastive sample pairs~\cite{chen2022we}, which in turn necessitates significant memory and extended pre-training times. To overcome this limitation, some recent approaches, such as the one utilized in MoCo~\cite{he2020momentum, chen2020improved}, have proposed sampling these pairs from a memory bank as an alternative to the approach employed in SimCLR~\cite{chen2020simple}. The latter method, which samples pairs from the current minibatch, results in higher memory consumption. The resource-intensive nature of contrastive learning for SSL has prompted researchers to explore alternative approaches, such as clustering (see supplementary material), distillation, and information maximization-based methods.

\paragraph{Clustering for SSL.} Clustering-based methods aim to extract useful representations by grouping data samples based on a similarity measure~\cite{caron2020unsupervised, bautista2016cliquecnn, yang2016joint, xie2016unsupervised, huang2019unsupervised, zhuang2019local, caron2019unsupervised, asano2019self, yan2020clusterfit}. For instance, DeepCluster~\cite{caron2020unsupervised} employs $k$-means clustering of representations from previous iterations as pseudo-labels for new representations, though this approach entails an expensive clustering phase performed asynchronously, making it challenging to scale up. Furthermore, clustering approaches can be seen as a form of contrastive learning at the cluster level, which still necessitates a substantial number of negative comparisons to perform effectively.

\paragraph{Distillation for SSL.} In contrast to using negative samples to prevent representation collapse, distillation-based approaches such as BYOL~\cite{richemond2020byol, richemond2020byol}, SimSiam~\cite{chen2021exploring}, and OBoW~\cite{gidaris2020learning} employ \textit{architectural techniques inspired by knowledge distillation.} These methods predominantly rely on a student-teacher network framework, in which the network's weights are either a moving average of the student network's weights~\cite{grill2020bootstrap} or are shared with the student network but with gradient updates stopped at the teacher network~\cite{chen2021exploring}. Although these methods can learn valuable representations, there is no clear evidence of how they prevent representation collapse, unlike contrastive learning approaches. Alternatively, in methods like OBoW~\cite{gidaris2020learning}, images can be represented as bags of words using a dictionary of visual features, which can be obtained through offline or online clustering.

\paragraph{Information Maximization for SSL.} These methods aim to prevent information collapse by \textit{maximizing the information in the embedding space}. Key works in this direction include Barlow-Twins~\cite{zbontar2021barlow}, Whitening-MSE~\cite{ermolov2021whitening}, VICReg~\cite{bardes2021vicreg}, and VicRegL~\cite{bardes2022vicregl}. In Barlow Twins, the loss term endeavors to make the normalized cross-correlation matrix of the embedding vectors from the two branches close to the identity. In Whitening-MSE, an additional module transforms the embeddings into the eigenspace of their covariance matrix, ensuring the obtained vectors are uniformly distributed on the unit sphere. VICReg extends the feature decorrelation concept from Barlow Twins by introducing two regularization terms: one to maintain the variance of each embedding dimension above a threshold and another to decorrelate each pair of variables. VICRegL~\cite{bardes2022vicregl} extends this idea further by proposing to learn both local and global features simultaneously. These infomax-based methods aim to produce embedding variables that are decorrelated, thus preventing informational collapse, as all variables are normalized over a batch, eliminating the incentive for them to shrink or expand. However, based on our experiments, we observe that while the aforementioned infomax-based methods avoid feature collapse with batch normalization, they tend to overfit or excessively focus on feature representations during pre-training, particularly on small and medium-scale datasets. This differs from contrastive learning methods and distillation-based methods, potentially leading to degraded downstream performance with extended pre-training. This might be attributed to the lack of actual interaction between samples, in contrast to the contrastive learning approaches. In order to overcome the above issue, we introduce another regularization on top of the Barlow Twins by introducing mixed samples into the SSL process where we assume that linear interpolated images will result in linearly interpolated embeddings. This simple trick can potentially avoid the feature memorization issue of infomax-based methods, particularly with Barlow Twins.

\paragraph{Mutual Sample Augmentations for SSL.} As previously mentioned, many SSL frameworks generate multiple views of each image during training (common in contrastive and info-max methods), and aim to teach the model that the embeddings of views from the same image should be as similar as possible. These views are often created using augmentations, which lead the model to become invariant to specific augmentations~\cite{purushwalkam2020demystifying, xiao2020should, zbontar2021barlow}.
Typically, these augmentations are applied on a per-sample basis. In contrast, supervised learning has demonstrated the effectiveness and generalization capabilities of augmentations that involve multiple samples~\cite{rebuffi2021data}, such as MixUp~\cite{zhang2017mixup} and CutMix~\cite{yun2019cutmix}. However, one primary reason for their limited adoption in SSL is the challenge of incorporating them into self-supervised loss formulations. There are very few works that attempted to use augmentations involving multiple images for SSL, such as MixCo~\cite{kim2020mixco}, i-Mix~\cite{lee2020imix}, Un-Mix~\cite{shen2022unmix}, Hard Negative Mixing~\cite{kalantidis2020hard}, and MNN~\cite{peng2023mnn}. Most of these methods are introduced for contrastive learning or clustering based approaches. \textit{In contrast, our proposed Mixed Barlow Twins formulation can be seen as integrating mutual sample augmentations into InfoMax-based SSL learning pipeline and has shown significant improvements in low- and medium-data regimes.}

\section{Proposed Method}
\label{sec:method}
\begin{figure}[tb]
    \centering
    \includegraphics[clip, trim=1cm 0 1.3cm 0, width=\linewidth]{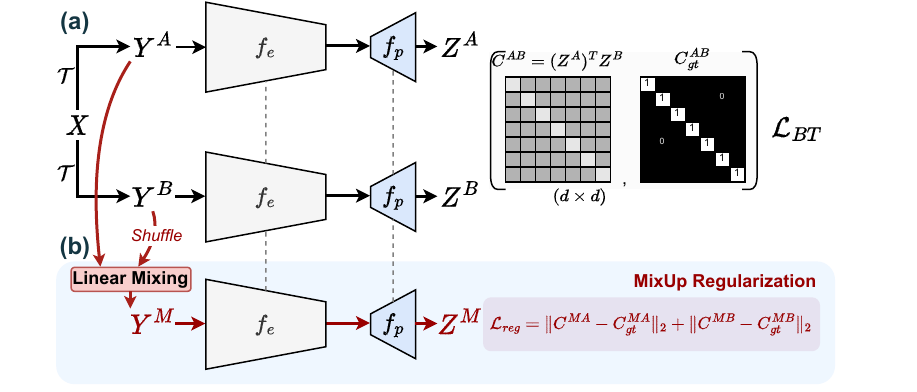}
    \caption{Schematic of the proposed Mixed Barlow Twins. (a) Original Barlow Twins Algorithm~\cite{zbontar2021barlow}. (b) Proposed MixUp regularization technique to prevent Barlow Twins from overfitting and to further enhance the representation quality.}
    \label{fig:method-bt-mixbt}
\end{figure}

\subsection{Overview of Barlow Twins}
Barlow Twins~\cite{zbontar2021barlow} adopts a Siamese network architecture consisting of two identical branches as shown in Figure \ref{fig:method-bt-mixbt}. Each branch processes a distinct view of the same input image $X$, denoted as $Y^A$ and $Y^B$. These views are generated by applying a series of random augmentations $\mathcal{T}$ to the original sample $X$. Augmentations include operations like random cropping, rotation, and color perturbations. Subsequently, $Y^A$ and $Y^B$ are forwarded through the encoder and the projector to obtain normalized embeddings $Z^A$ and $Z^B$ (centered along the batch dimension). These embeddings are then used to compute the Barlow Twins objective $\mathcal{L}_{BT}$ based on the cross-correlation matrix $C$ computed between $Z^A$ and $Z^B$ along the batch dimension as follows:
\begin{equation}
    \label{eq:loss_bt}
    C_{ij} \triangleq \sum_b \frac{ z^A_{b,i} \hspace{2pt} z^B_{b,j} }{\sqrt{\sum_{b}(z_{b,i}^A)^2 (z_{b,j}^B)^2}},
\end{equation}
where $b$ indexes batch samples, and $i$ and $j$ index the vector dimension of the embeddings. The Barlow Twins objective is defined on $C$ and consists of two terms:
\begin{enumerate}
    \item \textbf{Invariance Term:} The first term aims to equate the diagonal elements of $C$ to 1. This enforces invariance in the embeddings, making them resilient to the applied distortions. Mathematically, it is expressed as $\sum_{i} \left( 1- C_{ii} \right)^2$.
    \item \textbf{Redundancy Reduction Term:} The second term of the objective strives to equate the off-diagonal elements of $C$ to 0. This operation decorrelates different vector components of the embedding, reducing redundancy. The term is represented as $\lambda_{BT} \sum_{i} \sum_{j \neq i} C_{ij}^2$, where $\lambda_{BT}$ is a hyperparamter that controls the balance between invariance loss and redundacy reduction loss.
\end{enumerate}
The combination of these terms in Barlow Twins objective ensures that the embeddings are both invariant to distortions and exhibit reduced redundancy, resulting in highly informative and diverse feature representations.

\input{figs/method-overfitting} 
\subsection{Overfitting Issue of Barlow Twins} 
\label{sec:method-overfitting-issue}
While Barlow Twins algorithm boasts a straightforward design and a capacity to learn valuable representations for downstream tasks, we have observed a critical phenomenon: \textit{increasing the dimensionality of the embedding space $d$ can result in the production of lower-quality representations but also in the risk of overfitting}. This behavior may manifest as the network endeavors to minimize invariance and redundancy by memorizing individual samples. This issue, somewhat surprising, was neither acknowledged nor reported in the original Barlow Twins study, which primarily emphasized linear evaluation results conducted on the extensive ImageNet dataset.

However, our experiments conducted on smaller to medium-sized datasets, such as CIFAR-10, CIFAR-100, TinyImageNet, and STL-10, provide compelling evidence of the Barlow Twins's susceptibility to overfitting and the generation of suboptimal representations with the expansion of embedding dimensionality. To underscore this concern, we executed \knn evaluation and monitored the top-1 accuracy at five-epoch intervals throughout 1000 epochs of Barlow Twins training.
Based on the experimental results presented in Figure \ref{method-overfitting-bt}, a discernible trend emerges. Across all four datasets, increasing the embedding dimension initially results in improved top-1 accuracy. However, as pre-training extends, the utility of the learned representations diminishes, often starting to deteriorate around the 400-600 iteration mark. Consequently, employing representations obtained around this interval may prove more advantageous for the downstream tasks compared to relying on representations acquired after extensive pre-training.

\subsection{What Causes the Overfitting?}
The observed overfitting issue of Barlow Twins can be attributed to its unique loss formulation. To better understand this, let's compare the Barlow Twins' loss (Equation \ref{eq:loss_bt}) with the InfoNCE loss commonly used in contrastive SSL~\cite{chen2020simple, liu2021self}:
\begin{align*}
    \small
    \mathcal{L}_{\text{InfoNCE}} \triangleq - \underbrace{\sum_b \frac{\langle z^A_{b}, 
z^B_{b} \rangle_i}{\tau\norm{z^A_{b}}_2\norm{z^B_{b}}_2}}_\text{similarity term}\\
    \small
    + \underbrace{\sum_{b} \log\left( \sum_{b' \neq b} \exp \left(\frac{\langle z^A_{b}, 
z^B_{b'} \rangle_i}         {\tau\norm{z^A_{b}}_2\norm{z^B_{b'}}_2}\right)\right)}_\text{contrastive term},
\end{align*}
where $z^A$ and $z^B$ are the twin network outputs, $b$ indexes the sample in a batch, $i$ indexes the vector component of the output, and $\tau$ is a positive constant called temperature. As we can observe, the InfoNCE loss aims to maximize the variability among embeddings by increasing the pairwise distance between \textit{all sample pairs}. In contrast, Barlow Twins loss operates differently. It focuses on decorrelating the components of the embedding vectors rather than emphasizing the distance between samples within a batch. The distinction lies in Barlow Twins' approach, which results in \textit{no or less interaction} between the samples in a given batch. As the embedding dimension grows, it leads to a considerable increase in the total number of trainable parameters. This expansion can potentially lead to overfitting, where the optimization process predominantly involves memorizing the samples rather than substantially improving the quality of the embeddings. This deviation from encouraging sample variability to focus on decorrelation might be a contributing factor to the observed overfitting in Barlow Twins. 

\subsection{Guarding Barlow Twins from Overfitting}
Motivated by the aforementioned issue observed in Barlow Twins (in Figure \ref{method-overfitting-bt}), we now explore a potential solution to address it. While we recognize the advantages that the original Barlow Twins algorithm brings to SSL compared to contrastive learning approaches, our goal is to mitigate this issue through a simple modification.

To achieve this, we propose to incorporate an additional regularization term $\mathcal{L}_{reg}$ on top of Barlow Twins loss function $\mathcal{L}_{BT}$, \textit{which promotes interaction between samples} in the batch. This new regularization term $\mathcal{L}_{reg}$ is inspired by mixup regularization~\cite{zhang2017mixup} in supervised learning but adapted to the context of SSL, aligning with Barlow Twins loss formulation.

As depicted in Figure \ref{fig:method-bt-mixbt}, our approach involves promoting interaction between the samples by creating mixed samples from $Y^A$ and $Y^B$ through linear interpolation, obtaining the embeddings of the mixed samples from the network, and formulating the regularization loss by assuming network produces linearly interpolated embeddings for it. The following section provides a detailed explanation of the proposed Mixed Barlow Twins approach.

The proposed Mixed Barlow Twins first generates a batch of mixed samples denoted as $Y^M$ by linearly interpolating between $Y^A$ and $Y^B$. Since both $Y^A$ and $Y^B$ consist of different views from the same images, we first shuffle $Y^B$ to ensure that the linear interpolation predominantly involves different images. This process can be mathematically expressed as follows:
\begin{align}
    \label{eq:mix}
    Y^s &= \mathtt{Shuffle}(X^B),\\
    Y^M &= \lambda Y^A + (1-\lambda)Y^s,
\end{align}
where, $Y^S$ represents the shuffled batch of images obtained by shuffling images in $Y^B$ using a randomly determined shuffling order denoted as $\mathtt{Shuffle}(\cdot)$, and $\lambda$ is the interpolation ratio between $Y^A$ and $Y^S$, sampled from a Beta distribution: $\lambda \sim Beta(\alpha, \alpha)$~\cite{johnson1995continuous, zhang2017mixup}. In all of our experiments, we use $\alpha=1.0$ unless stated otherwise.

Next, we feed $Y^M$ through the encoder and projector to obtain their \textit{normalized} embeddings centered along the batch dimension:
\begin{equation}
    Z^M = f_{e+p} (Z^M),
\end{equation}
where $f_{e+p}(\cdot)$ denotes the network. Subsequently, we compute the cross-correlation matrices between the mixed embeddings $Z^M$ and the unmixed embeddings $Z^A$ and $Z^B$ along the batch dimension:
\begin{align}
    C^{MA} &= (Z^M)^TZ^A,\\
    C^{MB} &= (Z^M)^TZ^B,
\end{align}
where, $T$ denotes matrix transpose, $C^{MA}$ and $C^{MB} \in $ $\mathbb{R}^{d \times d}$, representing the cross-correlation between the mixed and the original embeddings. Assuming that ``\textit{linear interpolation in the input RGB space results in linearly interpolated features in the embedding space},'' we can determine the ground-truth cross-correlation matrices $C^{MA}_{gt}$ and $C^{MB}_{gt}$ for creating the regularization term:
\begin{align}
    C^{MA}_{gt} &= (Z^M)^TZ^A,\\
                &= (\lambda Z^A + (1-\lambda)Z^S)^T Z^A,\\
                \label{eq: C_gt_A}
                &= \lambda (Z^A)^TZ^A + (1-\lambda)\mathtt{Shuffle}^*(Z^B)^TZ^A.
\end{align}
Similarly,
\begin{equation}
    \label{eq: C_gt_B}
    C^{MB}_{gt} = \lambda (Z^A)^TZ^B + (1-\lambda)\mathtt{Shuffle}^*(Z^B)^TZ^B.
\end{equation}
Here, $\mathtt{Shuffle^*}$ denotes shuffling embeddings in the same order as in Equation (\ref{eq:mix}). The mixup-based regularization loss is designed to align $C^{MA}$ and $C^{MB}$ with their respective ground-truth cross-correlation matrices, $C^{MA}_{gt}$ and $C^{MB}_{gt}$. To achieve this, we employ a simple $L_2$ loss:
\begin{equation}
    \mathcal{L}_{reg} = \frac{\lambda_{BT}}{2} \left( \| C^{MA}-C^{MA}_{gt} \|_2 + \| C^{MB}-C^{MB}_{gt} \|_2 \right)
\end{equation}

The final loss $\mathcal{L}$ of the proposed Mixed Barlow Twins can then be expressed as:
\begin{equation}
    \mathcal{L} = \mathcal{L}_{BT} + \lambda_{reg} \mathcal{L}_{reg},
\end{equation}
where $\lambda_{reg}$ controls the balance between $\mathcal{L}_{BT}$ and $\mathcal{L}_{reg}$. 
As can be observed, the proposed mixup regularization \textit{improves the interaction between the samples} in the batch and \textit{introduces an infinite set of synthetic samples} into the pre-training process, which is demonstrated in the following section, leads to a reduction in feature overfitting and considerable improvements in the downstream performance.

\section{Experimental Results}
\label{sec:results}
\input{figs/results-bt-mixup}
\paragraph{Datasets.} We conduct experiments on datasets of varying scales to demonstrate the effectiveness of the proposed-mixup regularization in addressing the overfitting issue of Barlow Twins. We employ five datasets: CIFAR-10~\cite{krizhevsky2009learning}, CIFAR-100~\cite{krizhevsky2009learning}, TinyImageNet~\cite{le2015tiny}, STL-10~\cite{coates2011analysis}, and ImageNet-1k~\cite{deng2009imagenet}.
The CIFAR-10 and CIFAR-100 datasets consist of 50,000 training images and 10,000 test images, each with a size of $32\times32$ pixels. In contrast, TinyImageNet, which is a subset of the full ImageNet, comprises of 100,000 training images, 10,000 test images, and 200 classes, with images of dimension $64\times64$ pixels. Compared to CIFAR-10 and CIFAR-100, TinyImageNet is considered large due to its increased number of images and classes.
The STL-10 dataset is specifically designed for SSL and differs from the other three datasets. It includes a separate unlabeled dataset with 100,000 images, 5,000 training images, and 8,000 test images, and distributed across 10 classes. The unlabeled examples are drawn from a similar but broader image distribution, making it an ideal choice for SSL.
ImageNet~\cite{deng2009imagenet}, renowned as one of the largest image classification benchmarks, contains 1.2 million training images and 50,000 validation images. In our experiments, we employ the training set without labels for SSL training on all five datasets. However, for the STL-10 dataset, we incorporate additional unlabeled data into the SSL process.

\paragraph{Experimental Setup.} We employ ResNet-18 and ResNet-50~\cite{he2016deep} as the backbone architectures. In the case of SSL with CIFAR-10, CIFAR-100, TinyImageNet, and STL-10 datasets, we utilize the Adam~\cite{KingBa15} optimizer with a batch size of 256, cosine annealing learning rate scheduler~\cite{loshchilov2016sgdr} with linear warm up~\cite{loshchilovstochastic}, and a weight decay~\cite{loshchilov2018decoupled} of 1e-6 for pre-training. The pre-training phase spans 1000 epochs although we sometimes pre-trained for 2000 epochs on small datasets. We perform grid hyperparamter tuning on embedding dimension $d$ for values 128, 1024, 2048, and 4096,  $\lambda_{BT}$  for values 0.0078125 and 1/$d$~\cite{tsai2021note}, and $\lambda_{reg}$ with values $1 \times, 2 \times, 3 \times, 4 \times, 5 \times  \lambda_{BT}$. 
To assess the performance of the pre-trained models, we conduct \knn evaluation~\cite{fix1989discriminatory, cover1967nearest} on the test set and linear probing. \knn evaluation involves utilizing normalized features from the encoder $f_e$. We set $k=200$~\cite{ermolov2021whitening}. For linear probing, we employ the Adam optimizer with a batch size of 512, exponential learning scheduler~\cite{li2019exponential}, and a weight decay of 1e-6. We use single NVIDIA RTX A5000 GPU for all experiments on CIFAR-10, CIFAR-100, TinyImageNet, and STL-10 datasets. 

For the experiments on ImageNet, we conducted experiments using the ResNet50 backbone. We employed the LARS~\cite{you2017large} optimizer with a batch size of 1024 distributed across eight(8) NVIDIA RTX A5000 GPUs, with base learning rates of 0.2 for weights and 0.0048 for biases, respectively. Both learning rates were linearly ramped from 0 up to their base values over 10 epochs and then gradually decayed with a cosine schedule over the remaining epochs until they reached 0.001 times their base rate. All ImageNet experiments were trained for 300 epochs with an embedding dimension of 8192. We used weight decay of 1e-6 and set $\lambda_{BT}$ to 0.0051. The projector consisted of MLPs with an embedding dimension of 8192-8192-8192. Following the original implementation, cross-correlation matrices computed on each GPU were combined using the \texttt{all\_reduce} operation before computing the Barlow Twins loss $\mathcal{L}_{BT}$ and the MixUp regularization loss $\mathcal{L}_{reg}$. For linear probing, we freeze the ResNet50 backbone and train a classifier for 100 epochs with a batch size of 256 distributed across 8 GPUs. We utilize a cosine annealing learning rate scheduler with a base learning rate of 0.3. We use the Stochastic Gradient Descent (SGD) optimizer with a momentum of 0.9 and weight decay of 1e-6. During the linear probing, random resized crops of 224x224 are used, followed by random horizontal flips as the only augmentations applied. During testing phase, images are resized to 256 and then center-cropped to 224.

Our implementation of Mixed Barlow Twins is based on the original Barlow Twins implementation \footnote{\scriptsize \url{https://github.com/facebookresearch/barlowtwins}} for experiments on ImageNet and Barlow Twins HSIC~\cite{tsai2021note}\footnote{\scriptsize \url{https://github.com/yaohungt/Barlow-Twins-HSIC}} with modifications mentioned above for other datasets. For SOTA methods (SimCLR~\cite{chen2020simple}, BYOL~\cite{richemond2020byol}, and WiteningMSE~\cite{ermolov2021whitening}), we closely follow their original implementations for ImageNet and the implementations by \cite{ermolov2021whitening}\footnote{\scriptsize \url{https://github.com/htdt/self-supervised}} for other datasets. 

\begin{table*}[tbh!]
\begin{center}
\caption{Comparing Mixed Barlow Twins with SOTA methods using \textbf{ResNet-50}~\cite{he2016deep} backbone.}
\label{table.sota-resnet50}
{\renewcommand{\arraystretch}{1.25}
\begin{tabular}{l|c|r r|r r|r r|r r}
\toprule
Method & \multirow{2}{*}{Epochs} & \multicolumn{2}{c |}{CIFAR-10} & \multicolumn{2}{c |}{CIFAR-100} & \multicolumn{2}{c |}{TinyImageNet} & \multicolumn{2}{c}{STL-10}\\
 & & \knn & linear & \knn & linear & \knn & linear & \knn & linear\\
\midrule

SimCLR \cite{chen2020simple} 
& 1000
& 88.94 & 91.85 
& 57.55 & 68.37
& 29.62  & 46.34
& 85.14 & 89.26\\

BYOL \cite{richemond2020byol}
& 1000
& 90.11 & 92.76
& 61.25 & 69.59
& 25.78  & 41.02
& 86.94 & 91.18\\


Barlow Twins \cite{zbontar2021barlow} 
& 1000
& 85.92 & 90.88
& 57.93 & 66.15
& 37.66 & 46.86
& 84.78 & 87.93 \\


\hdashline
Mixed Barlow Twins (ours) 
& 1000
& \bf 91.14 & \bf 93.48  
& \bf 61.71 & \bf 71.98
& \bf 40.52 & \bf 50.59
& \bf 87.55 & 91.10\\

Mixed Barlow Twins (ours) 
& 2000
& \bf 91.39 & \bf \bf 93.89
& \bf 64.32 & \bf 72.51
& \bf 42.21 & \bf 51.84
& \bf 87.79 & \bf 91.70 \\

\bottomrule

\end{tabular}}
\end{center}
\end{table*}
\begin{table*}[t]
\begin{center}
\caption{Comparing Mixed Barlow Twins with SOTA methods using \textbf{ResNet-18}~\cite{he2016deep} backbone.}
\label{table.sota-resnet18}
{\renewcommand{\arraystretch}{1.25}
\begin{tabular}{l|c|r r|r r|r r|r r}
\toprule
Method & \multirow{2}{*}{Epochs} &\multicolumn{2}{c |}{CIFAR-10} & \multicolumn{2}{c |}{CIFAR-100} & \multicolumn{2}{c |}{TinyImageNet} & \multicolumn{2}{c}{STL-10} \\
 & & \knn & linear & \knn & linear & \knn & linear & \knn & linear \\
\midrule

SimCLR \cite{chen2020simple} 
& 1000
& 88.42 & 91.80
& 56.56 & 66.83
& 32.86  & 48.84
& 85.68 & 90.51\\

BYOL \cite{richemond2020byol}
& 1000
& 89.45 & 91.73
& 56.82 & 66.60
& 36.24 & 51.00
& 88.64 & \bf 91.99\\

W-MSE \cite{ermolov2021whitening}
& 1000
& 89.69 & 91.55 
& 56.69 & 66.10
& 34.16  & 48.20
& 87.10 & 90.36\\


Barlow Twins~\cite{zbontar2021barlow} 
& 1000
& 86.66 & 87.76
& 55.94 & 61.64
& 33.60 & 41.80
& 84.23  & 88.21\\

\hdashline 

Mixed Barlow Twins (ours) 
& 1000
& \bf 89.91 & \bf 91.99
& \bf 61.12 & \bf 68.55
& \bf 37.52 & \bf 51.48
& 88.23 & 91.04\\

Mixed Barlow Twins (ours) 
& 2000
& \bf 90.52 & \bf 92.58
& \bf 61.25 & \bf 69.31
& \bf 38.11 & \bf 51.67 
& \bf 88.94 & 91.02\\

\bottomrule

\end{tabular}}
\end{center}
\end{table*}
\paragraph{Effect of MixUp Regularization: Comparing \knn Evaluation Results with the ResNet50 Backbone.}
After demonstrating the overfitting issue of the Barlow Twins algorithm in Section \ref{sec:method-overfitting-issue}, we now delve into the \knn evaluation results to understand how they evolve throughout SSL training when mixup regularization $\mathcal{L}_{reg}$ is incorporated. As illustrated in Figure \ref{fig:knn-results-bt-mixup-top1}, where solid lines represent SSL training with $\mathcal{L}_{BT}+\mathcal{L}_{reg}$ and dashed lines correspond to $\mathcal{L}_{BT}$ alone, it becomes evident that integrating mixup regularization into Barlow Twins training always leads to improved \knn accuracy and reduced overfitting across all four datasets. Across all four embedding dimensions considered, the proposed Mixed Barlow Twins eliminates the overfitting issue, leading to improved top-1 accuracy throughout SSL training. The most notable improvement is seen with an embedding dimension of $d=1024$, where we observe a remarkable \textbf{+5.2\%} increase over the original Barlow Twins method with mixup regularization (91.14\% \textit{vs.} 85.93\%) on \textbf{CIFAR-10}. Similarly, in the case of \textbf{CIFAR-100}, we witness a similar trend. Incorporating mixup regularization yields better \knn evaluation performance. When comparing the best performance, achieved with an embedding dimension of $d=1024$, we note a substantial +3.78\% improvement for Mixed Barlow Twins over the Barlow Twins (61.71\% \textit{vs.} 57.93\%). When considering \textbf{TinyImageNet}, where the best performance is achieved with a embedding dimension of $d=1024$, which is comparatively smaller than the optimal results for CIFAR-10 and CIFAR-100. In the best-case scenario, we observe a \textbf{+2.86\%} improvement with mixup regularization (40.52\% \textit{vs.} 37.66\%). Experiments with the \textbf{STL-10} dataset presents similar observations. We notice a \textbf{+2.77\%} improvement (87.55\% \textit{vs.} 84.78\%)  for the best-case scenario where embedding dimension is $d=1024$. 

In summary, the results obtained across all four datasets consistently demonstrate that adding the proposed mixup-based regularization on top of the original Barlow Twins loss mitigates network overfitting and fosters the learning of superior feature representations that prove beneficial for downstream applications. 

\paragraph{Comparison with SOTA methods using ResNet-18 and ResNet-50.} Table \ref{table.sota-resnet50} and \ref{table.sota-resnet18} compare the results of our Mixed Barlow Twins with Barlow Twins and other SOTA approaches using ResNet-50 and ResNet-18, respectively. We present both \knn evaluation and linear probing results (top-1). When looking at the \knn results with the ResNet-18 backbone, we can observe considerable improvements that Mixed Barlow Twins brings over Barlow Twins and other SOTA methods on all four datasets. More importantly, Mixed Barlow Twins achieves new SOTA results with  \textbf{+1.28\%} (91.39 \textit{vs.} 90.11), \textbf{+3.07\%} (64.32 \textit{vs.} 61.25), \textbf{+4.45\%} (42.21 \textit{vs.} 37.66), and  \textbf{+0.85\%} (87.79 \textit{vs.} 86.94) with \knn evaluation and \textbf{+1,13\%}, \textbf{+2.92\%}, \textbf{+4.98\%}, and \textbf{+0.52\%} with linear evaluation on CIFAR-10, CIFAR-100, TinyImageNet, and STL-10 datasets, respectively, using the  \textbf{ResNet-50} backbone. As shown in Table \ref{table.sota-resnet18}, similar performance improvements over Barlow Twins and other SOTA methods are observed using the ResNet-18 backbone, except it achieves second best results on the STL-10 dataset under linear evaluation. 

\begin{figure}[htb]
    \centering
    \label{fig:in-training-loss}
    \begin{subfigure}[b]{\linewidth}
        \centering
        \includegraphics[width=\linewidth]{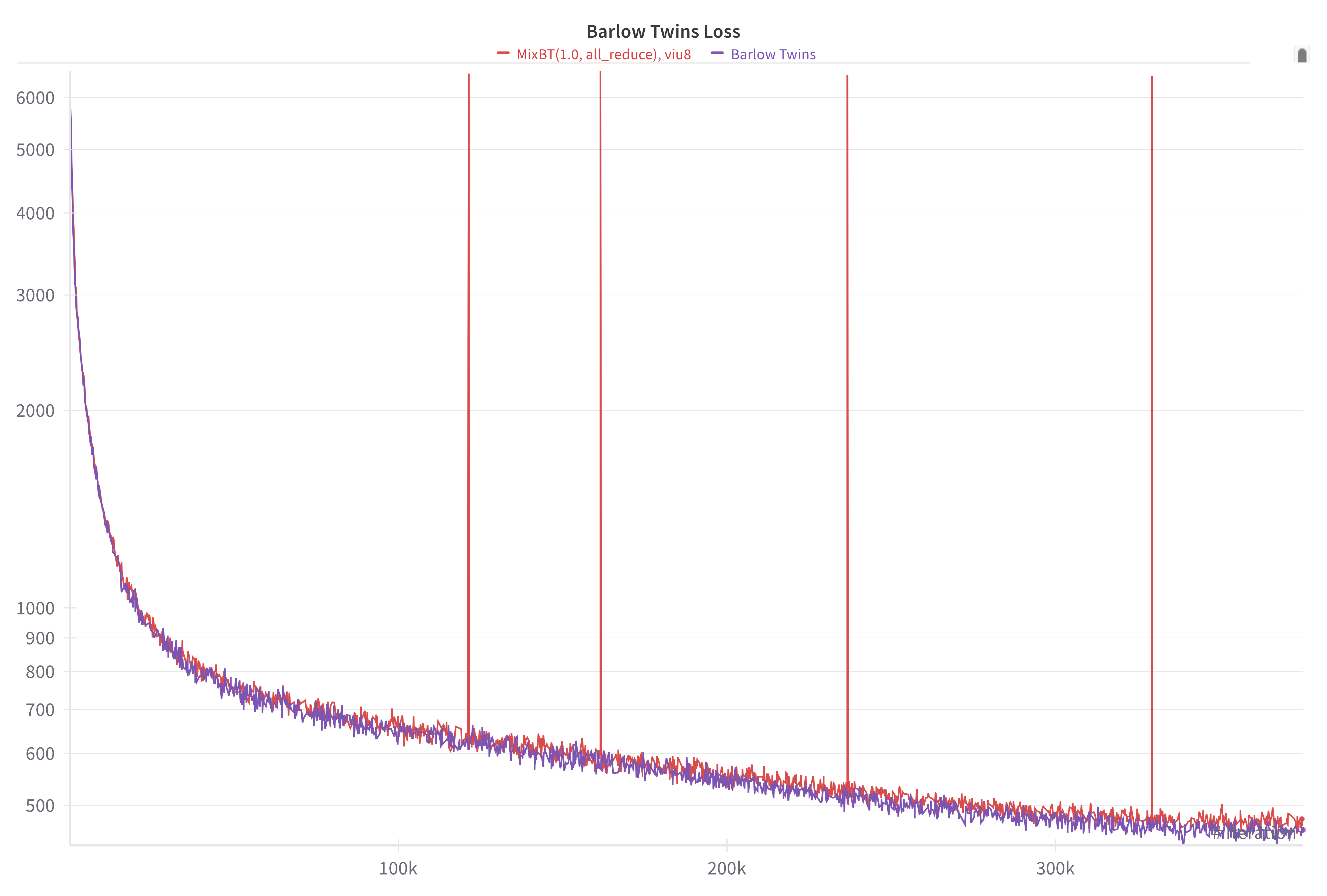}
        \caption{Barlow Twins Loss $\mathcal{L}{BT}$ for Mixed Barlow Twins Training (in red) and Barlow Twins Training (in purple).}
        \label{fig:in-bt-loss}
    \end{subfigure}
    \hfill
    \begin{subfigure}[b]{\linewidth}
        \centering
        \includegraphics[width=\linewidth]{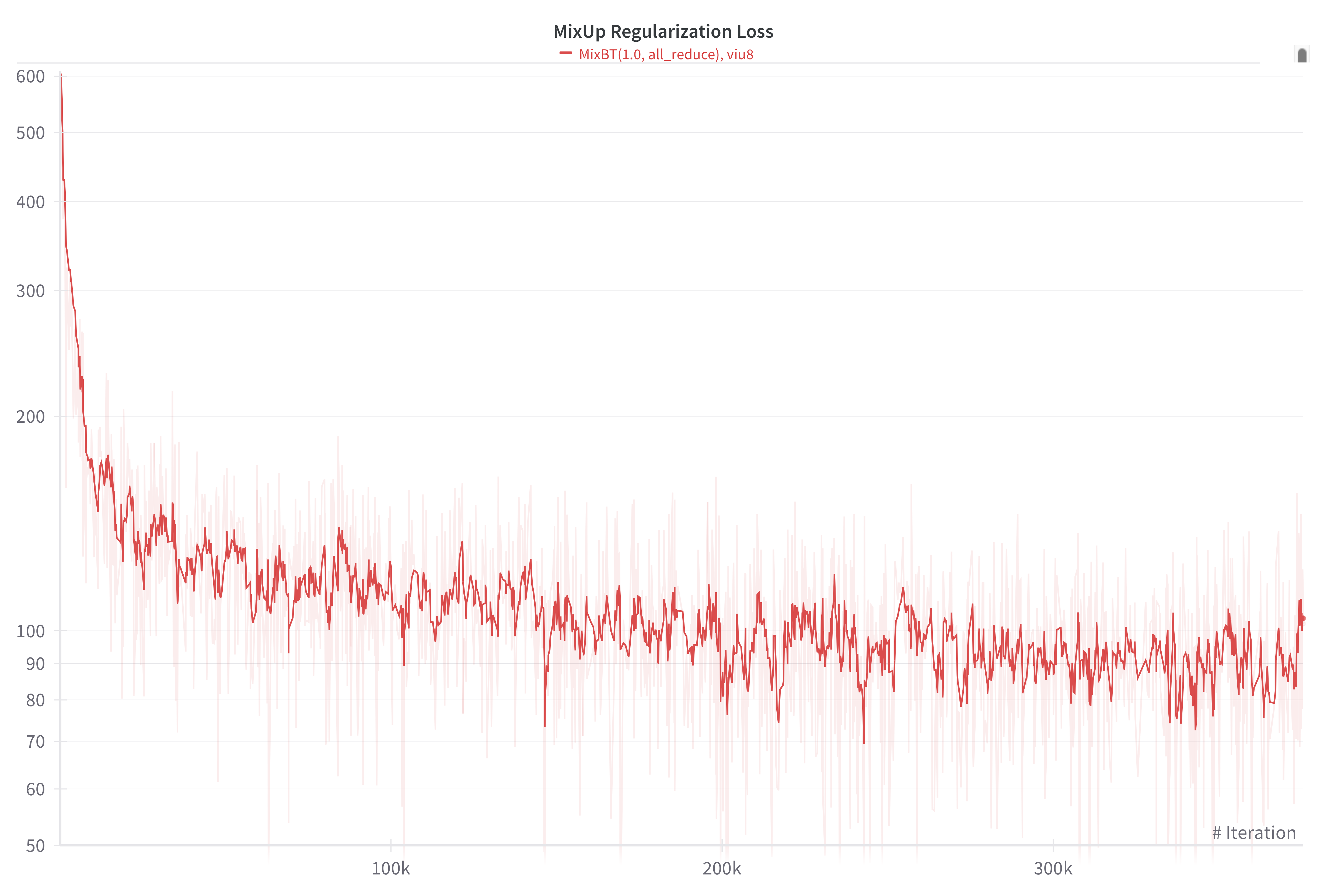}
        \caption{MixUp Regularization Loss $\mathcal{L}{reg}$ for Mixed Barlow Twins Training (in red). In this experiment, $\lambda_{reg}$ is set to 1.0.}
        \label{fig:in-mix-bt-loss}
    \end{subfigure}
    \caption{Convergence of Barlow Twins loss and mixup-based regularization loss during ImageNet training for 300 epochs. Barlow Twins training is shown in purple color and Mixed Barlow Twins training is shown in red color.}
\end{figure}

\paragraph{Results on ImageNet-1k.}
\begin{table}[tbh!]
\caption{Linear probing results (top-1 \%) on \textbf{ImageNet}~\cite{deng2009imagenet} with \textbf{ResNet50} backbone.}
\centering
\begin{tabular}{lp{0.9cm}p{1.4cm}}
    \toprule
     Method   & Epochs &  Acc. \%\\ 
    \midrule
    MoCo v2~\cite{chen2020improved}  & 200 & 67.5\\
    SimCLR~\cite{chen2020simple}  & 400 & 68.2\\
    BYOL~\cite{richemond2020byol}  & 300 & \textbf{72.3}\\
    VICReg~\cite{bardes2021vicreg}  & 300 & 71.5\\
    VICRegL~\cite{bardes2022vicregl} & 300 & 71.2\\
    Barlow Twins~\cite{zbontar2021barlow} & 300 & 71.3\\
    \hdashline
    \textbf{Ours:} & \\
    \hspace{5pt} Mixed Barlow Twins \tiny $( \mathcal{\lambda}_{reg}=0.0025)$ & 300 & 70.9\\
    \hspace{5pt} Mixed Barlow Twins \tiny $( \mathcal{\lambda}_{reg}=0.1)$ & 300 & 71.6\\
    \hspace{5pt} Mixed Barlow Twins \tiny $( \mathcal{\lambda}_{reg}=1.0)$ & 300 & \underline{72.2}\\
    \bottomrule
\end{tabular}
\label{tab:imagenet_result_resnet50}
\end{table} 
Before delving into the linear probing results, we provide some training statistics that compare the convergence of different loss terms during training. Figures \ref{fig:in-bt-loss} and \ref{fig:in-mix-bt-loss} illustrate the convergence of two loss terms: the Barlow Twins loss ($\mathcal{L}{BT}$) and the MixUp regularization loss ($\mathcal{L}{reg}$). These figures compare the default Barlow Twins training (shown in purple) with Mixed Barlow Twins training (shown in red), specifically for a regularization weight ($\lambda_{\text{reg}}$) value of 1.0. In the case of Mixed Barlow Twins training, we observe that the Barlow Twins loss ($\mathcal{L}_{BT}$) tends to converge to a slightly higher average value (on average) compared to the default Barlow Twins training. This difference in convergence is likely due to the influence of MixUp regularization. However, despite the higher Barlow Twins loss, the results of linear evaluation indicate that Mixed Barlow Twins training achieves a higher top-1 accuracy compared to its default counterpart.

Table \ref{tab:imagenet_result_resnet50} presents the linear evaluation results on ImageNet with the ResNet-50 backbone. It is evident that Mixed Barlow Twins achieves the second-best results compared to SOTA methods. Moreover, it significantly improves over Barlow Twins and its later counterparts, such as VICReg and VicRegL. This outcome indicates that introducing mixed samples into the SSL framework is beneficial even for large datasets.
\section{Further Discussion}
\label{sec:further_diss}
\input{figs/btloss-cofar10-1024}
\paragraph{Convergence of Different Loss Terms.} In this section, we examine how each of the loss terms converges during the SSL training for default Barlow Twins (i.e., $\mathcal{L}_{BT}$) and Mixed Barlow Twins training (i.e., $\mathcal{L}_{BT}$ and $\mathcal{L}_{reg}$). From Figure \ref{fig:loss-breakdown}, it is evident that, despite Mixed Barlow Twins demonstrating improved performance (see Figure \ref{fig:knn-results-bt-mixup-top1}), it converges to a higher Barlow Twins loss under MixUp regularization compared to the default Barlow Twins training. \textit{This observation underscores a key insight – achieving lower Barlow Twins loss during training does not inherently guarantee improved downstream performance.} In fact, the drive to excessively minimize the Barlow Twins loss may lead the network to memorize or overly fixate on the embeddings of samples, which can ultimately result in poor downstream performance. In contrast, incorporating mixed samples into the SSL training makes it challenging for the network to memorize samples, as it introduces a theoretically infinite variety of mixed samples into the SSL training process, effectively compelling the network to minimize the Barlow Twins loss by learning valuable high-level representations. These representations, in turn, prove beneficial for the downstream applications, ultimately leading to the enhanced performance. 

\paragraph{Effect of Mixup Regularization Coefficient $\mathcal{\lambda}_{reg}$.}
\input{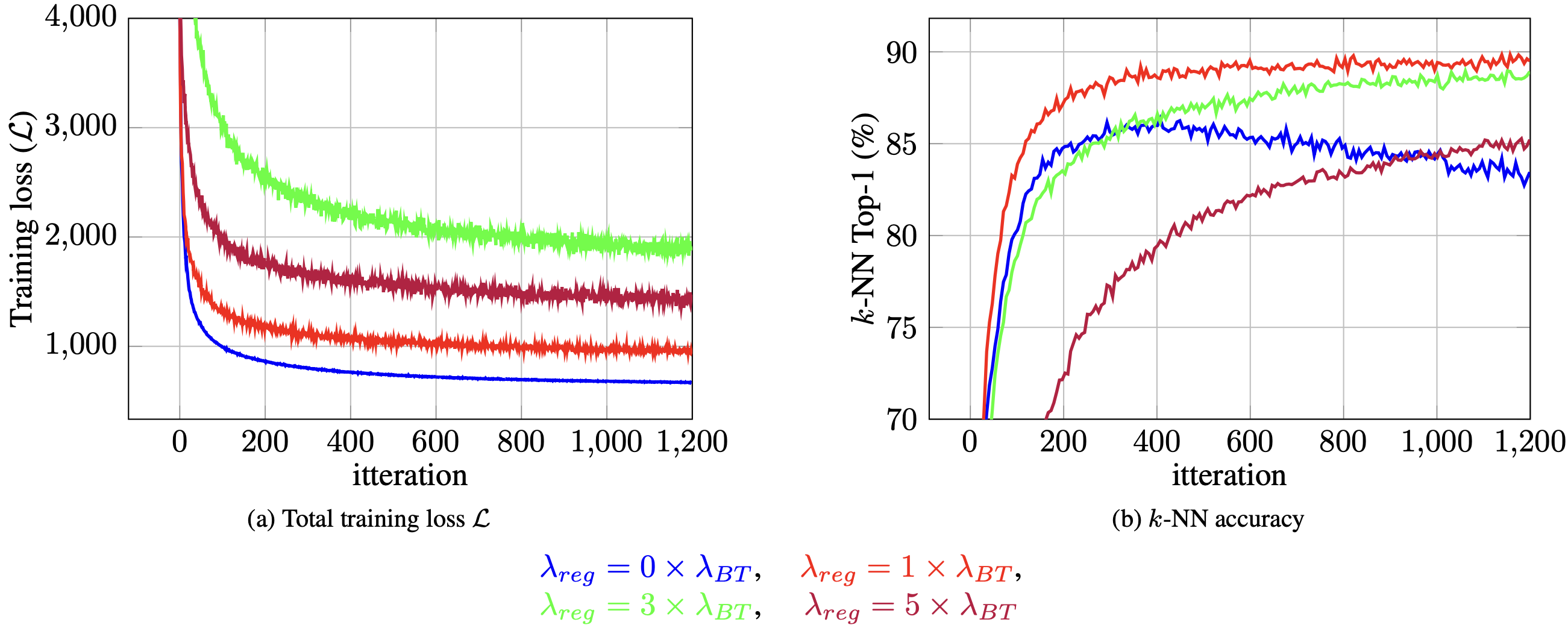}
In this section, we examine the impact of the regularization parameter $\lambda_{reg}$ on the downstream performance. As previously observed in Section \ref{sec:results}, the addition of mixup regularization to the original Barlow Twins loss significantly enhances the downstream performance. It is crucial to note that, like any other regularization term, selecting an appropriate coefficient that adequately balances the loss terms is essential. A substantial increase in the regularization coefficient noticeably prolongs the convergence time of the network. Figure \ref{fig:abl-mixup-coefficient} illustrates the variations in the total training loss $\mathcal{L}$ and the \knn evaluation accuracy throughout the SSL training for different mixup regularization coefficient values. The figure demonstrates that an escalation in the mixup regularization coefficient leads to slower convergence and reduced downstream performance. This phenomenon was consistently observed across the other three datasets (CIFAR-100, TinyImageNet, and STL-10). Based on our analysis, we found that setting $\lambda_{reg} = 1 \text{ to } 3 \times \lambda_{BT}$ serves as a useful rule of thumb.

\section{Conclusion}
\label{sec:conclusion}
We highlight some limitations of the popular Barlow Twins algorithm and propose a remedy in the form of Mixed Barlow Twins. Our experiments demonstrate that the integration of mixup-based regularization effectively mitigates feature overfitting and significantly enhances the downstream task performance. This novel approach enriches the Barlow Twins process by promoting the interaction between samples and introducing a myriad of synthetic samples. The results endorse the efficacy of Mixed Barlow Twins in addressing the drawbacks of the Barlow Twins approach, suggesting a promising direction in SSL.
\appendix
\maketitlesupplementary
\section{Transfer Learning}
\begin{table*}[htb]
\centering
\begin{tabular}{lccccccc}
    \toprule[0.01em]
    Model       & DTD & MNIST & FaMNIST & CUBirds & VGGFlower & TrafficSigns & Aircraft \\ \midrule
    
    \multicolumn{8}{l}{\bf CIFAR-10} \\
    Barlow Twins & 20.53 & 91.78 & 83.49 & 5.04 & 25.80 & 49.42 & 14.58\\
    Mixed Barlow Twins (ours) & \bf 34.04 & \bf 97.26 & \bf 87.84 & \bf 10.70 & \bf 56.48 & \bf 76.06 & \bf 31.13\\ \midrule

    \multicolumn{8}{l}{\bf CIFAR-100} \\
    Barlow Twins & 27.45 & 93.72 & 84.78 & 5.89 & 34.96 & 59.85 & 15.54\\
    Mixed Barlow Twins (ours) & \bf 37.23 & \bf 97.81 & \bf 88.03 & \bf 11.01 & \bf 64.04 & \bf 77.61 & \bf 31.23\\ \midrule

    \multicolumn{8}{l}{\bf STL-10} \\
    Barlow Twins & 45.10 & 96.34 & 85.28 & 11.53 & 68.03 & 66.80 & 34.65\\
    Mixed Barlow Twins (ours) & \bf 46.31 & \bf 97.31 & \bf 86.21 & \bf 12.27 & \bf 68.36 & \bf 69.73 & \bf 35.27\\

    \bottomrule    
\end{tabular}
\caption{Transfer learning results (linear classification accuracy) from CIFAR-10 $\rightarrow$ to \{DTD, MNIST, FaMNIST, CUBirds, VGGFlower, TrafficSigns\} with ResNet18 backbone. Both models are pre-trained for 1000 epochs on CIFAR-10.}
\label{tab:tf-resnet18-cifar10}
\end{table*}
In this section, we compare transfer learning capabilities of Mixed Barlow Twins with Barlow Twins.
\subsection{Transfer Learning Datasets}
For the transfer learning experiments, we consider seven datasets:
\begin{enumerate}
    \item \textbf{DTD}~\cite{dtd}: This texture database consists of 5640 images, organized into 47 categories inspired by human perception. Each category contains 120 images. The image sizes range from 300x300 to 640x640 pixels, and each image contains at least 90\% of the surface representing the category attribute.  
    \item \textbf{MNIST}~\cite{mnist}: MNIST is a dataset of handwritten digits, with a training set of 60,000 examples and a test set of 10,000 examples.
    \item \textbf{FashionMNIST}~\cite{fashion-mnist}: FashionMNIST is a dataset of Zalando's article images, consisting of a training set of 60,000 examples and a test set of 10,000 examples. Each example is a 28x28 grayscale image associated with a label from one of 10 classes.
    \item \textbf{CUBirds}~\cite{cu-birds}: CUBirds is a challenging image dataset annotated with 200 bird species, mostly North American. It contains 11,788 images categorized into 200 subcategories, with 5,994 images for training and 5,794 for testing. Each image has detailed annotations, including a subcategory label, 15 part locations, 312 binary attributes, and a bounding box.
    \item \textbf{VGGFlower}~\cite{vgg-flowers}: VGGFlower is a dataset with 102 categories, consisting of 102 flower categories commonly occurring in the United Kingdom. Each class contains between 40 and 258 images.
    \item \textbf{Traffic Signs}~\cite{traffic-signs}: This dataset is from the German Traffic Sign Detection Benchmark (GTSDB) and includes 900 training images (1360 x 800 pixels) containing only the traffic signs.
    \item \textbf{Aircraft}~\cite{aircraft}: Aircraft is a benchmark dataset for fine-grained visual categorization of aircraft. It contains 10,200 images of aircraft, with 100 images for each of 102 different aircraft model variants, most of which are airplanes. Each image is annotated with a tight bounding box and a hierarchical airplane model label.
\end{enumerate}

\subsection{Transfer Learning Results}
We present linear evaluation results in which a linear classifier is trained on top of the frozen ResNet backbone. The linear classifier is trained for 100 epochs using the SGD optimizer. Table \ref{tab:tf-resnet18-cifar10} displays the transfer learning results from CIFAR-10, CIFAR-100, and STL-10 to \{DTD, MNIST, FashionMNIST, CUBirds, VGGFlower, Traffic Signs, and Aircraft\}, with the best result in each block shown in bold. It can be observed that our Mixed Barlow Twins consistently yield significant improvements over Barlow Twins, demonstrating that mixup regularization not only enhances in-dataset performance but also boosts the transfer learning capabilities of the model.

\section{Pseudocode}
\begin{algorithm*}[t]
   \caption{PyTorch-style pseudocode for Mixed Barlow Twins with changes required on top of Barlow Twins highlighted in red. This pseudocode template is adapted from Barlow Twins.}
   \label{alg:mixed_barlow_twins}
   
    \definecolor{codeblue}{rgb}{0.25,0.5,0.5}
    \lstset{
      basicstyle=\fontsize{8pt}{8pt}\ttfamily\bfseries,
      commentstyle=\fontsize{8pt}{8pt}\color{codeblue},
      keywordstyle=\fontsize{8pt}{8pt},
    }
\begin{lstlisting}[language=python]
# f: encoder network
# lambda: weight on the off-diagonal terms
# lmbda_mixup: weight on the mixup regularization loss
# N: batch size
# D: dimensionality of the embeddings
#
# mm: matrix-matrix multiplication
# off_diagonal: off-diagonal elements of a matrix
# eye: identity matrix
# randperm: random permutation of integers
# beta: draw samples from a Beta distribution

for x in loader: # load a batch with N samples
    # two randomly augmented versions of x
    y_a, y_b = augment(x)
    
    # compute embeddings
    z_a = f(y_a) # NxD
    z_b = f(y_b) # NxD
    
    # normalize repr. along the batch dimension
    z_a_norm = (z_a - z_a.mean(0)) / z_a.std(0) # NxD
    z_b_norm = (z_b - z_b.mean(0)) / z_b.std(0) # NxD
    
    # cross-correlation matrix
    c = mm(z_a_norm.T, z_b_norm) / N # DxD
    
    # loss
    c_diff = (c - eye(D)).pow(2) # DxD
    # multiply off-diagonal elems of c_diff by lambda
    off_diagonal(c_diff).mul_(lambda)
    loss_bt = c_diff.sum()
\end{lstlisting}
\begin{lstlisting}[language=python, backgroundcolor=\color{pink}]
    ### MixUp Regularization (our contribution) ###
    # creating mixed samples: Eqn. (2)-(3)
    idxs = randperm(N)
    alpha = beta(1.0, 1.0)
    y_m = alpha * y_a + (1 - alpha) * y_b[idxs, :]

    # compute mixed sample embeddings: Eqn. (4) 
    z_m = f(y_m) # NxD
    z_m_norm = (z_m - z_m.mean(dim=0)) / z_m.std(dim=0) # NxD

    # cross-correlation matrices: Eqn. (5)-(6)
    cc_m_a = mm(z_m_norm.T, z_a_norm) / N  # DxD
    cc_m_b = mm(z_m_norm.T, z_b_norm) / N  # DxD

    # groud-truth cross-correlation matrices: Eqn. (9)-(10)
    cc_m_a_gt = alpha*mm(z_a_norm.T, z_a_norm)/N + (1-alpha)*mm(z_b_norm[idxs,:].T, z_a_norm)/N # DxD
    cc_m_b_gt = alpha*mm(z_a_norm.T, z_b_norm)/N + (1-alpha)*mm(z_b_norm[idxs,:].T, z_b_norm)/N # DxD

    # mixup regularization loss: Eqn. (11)
    loss_mix = lmbda_mixup*lmbda*((cc_m_a-cc_m_a_gt).pow_(2).sum() + (cc_m_b-cc_m_b_gt).pow_(2).sum())

\end{lstlisting}
\begin{lstlisting}[language=python]  
    # total loss
    loss = loss_bt + loss_mix

    # optimization step
    loss.backward()
    optimizer.step()
\end{lstlisting}
\end{algorithm*}
We present pseudocode for our Mixed Barlow Twins algorithm in Algorithm \ref{alg:mixed_barlow_twins}. Since our implementation is based on the default Barlow Twins implementation, we have highlighted the modifications required to be added to Barlow Twins in red color. As shown in the pseudocode, our method involves a few lines of code changes on top of the Barlow Twins algorithm.

\section{Default Hyperparameter Configuration for Mixed Barlow Twins Training}
In this section, we provide the default (i.e., the best) hyperparameter configuration used for pre-training on each dataset.
\subsection{CIFAR-10/CIFAR-100}
Table \ref{tab:mbt-resnet18-cifar10-cifar-100} presents the default configuration for the Mixed Barlow Twins experiments conducted on CIFAR-10 and CIFAR-100 datasets using ResNet-18 and ResNet-50 as backbones. We observed that among the choices of projector dimension $d$ (128, 1024, 2048, and 4096), a value of 1,024 consistently yielded the best results when combined with a $\lambda_{BT}$ value of 0.0078125 (chosen from the options 0.0078125 and $1/d$), particularly for CIFAR datasets. The optimal nearest neighbor (\knn) accuracy was achieved with a $\lambda_{reg}$ value of 4.0.

\begin{table}[H]
\centering
\caption{Default Hyperparameter Configuration for Mixed Barlow Twins Training on CIFAR-10 and CIFAR-100 datasets with ResNet-18 and ResNet-50 Backbones.}
\label{tab:mbt-resnet18-cifar10-cifar-100}
    \begin{tabular}{|c|c|}
    \hline
    \textbf{Key} & \textbf{Value} \\
    \hline
    batch size & 256 \\
    learning rate & 0.01 \\
    learning rate scheduler & ``cosine'' \\
    feature dim. $(d)$ & 1,024 \\
    $\lambda_{BT}$ & 0.0078125 \\
    $\lambda_{reg}$ & 4.0 \\
    \hline
    \end{tabular}
\end{table}

\subsection{TinyImageNet}
Table \ref{tab:mbt-resnet18-tiny} presents the default configuration for the Mixed Barlow Twins experiments conducted on TinyImageNet using ResNet-18 and ResNet-50 as backbones. We observed that among the choices of projector dimension $d$ (128, 1024, 2048, and 4096), a value of 1,024 consistently yielded the best results when combined with a $\lambda_{BT}$ value of 0.0009765 (chosen from the options 0.0078125 and $1/d$=0.0009765). The optimal nearest neighbor (\knn) accuracy was achieved with a $\lambda_{reg}$ value of 4.0.

\begin{table}[H]
\centering
\caption{Default hyperparameter configuration for Mixed Barlow Twins training on TinyImageNet with ResNet-18/ResNet-50 backbone.}
\label{tab:mbt-resnet18-tiny}
    \begin{tabular}{|c|c|}
    \hline
    \textbf{Key} & \textbf{Value} \\
    \hline
    batch size & 256 \\
    learning rate & 0.01 \\
    learning rate scheduler & ``cosine'' \\
    feature dim. $(d)$ & 1,024 \\
    $\lambda_{BT}$ & $1/d$ = 0.0009765 \\
    $\lambda_{reg}$ & 4.0 \\
    \hline
    \end{tabular}
\end{table}

\subsection{STL-10}
Table \ref{tab:mbt-resnet18-stl10} presents the default configuration for the Mixed Barlow Twins experiments conducted on TinyImageNet using ResNet-18 and ResNet-50 as backbones. We observed that among the choices of projector dimension $d$ (128, 1024, 2048, and 4096), a value of 1,024 consistently yielded the best results when combined with a $\lambda_{BT}$ value of 0.0078125 (chosen from the options 0.0078125 and $1/d$). The optimal nearest neighbor (\knn) accuracy was achieved with a $\lambda_{reg}$ value of 2.0.
\begin{table}[H]
\centering
\caption{Default hyperparameter configuration for Mixed Barlow Twins training on STL-10 with ResNet-18/ResNet50 backbone.}
\label{tab:mbt-resnet18-stl10}
    \begin{tabular}{|c|c|}
    \hline
    \textbf{Key} & \textbf{Value} \\
    \hline
    batch size & 256 \\
    learning rate & 0.01 \\
    learning rate scheduler & ``cosine'' \\
    feature dim. $(d)$ & 1,024 \\
    $\lambda_{BT}$ & 0.0078125 \\
    $\lambda_{reg}$ & 2.0 \\
    \hline
    \end{tabular}
\end{table}

\subsection{ImageNet}
Table \ref{tab:mbt-resnet18-in} presents the default configuration for the Mixed Barlow Twins experiments conducted on ImageNet using ResNet-50 as the backbone. We use the default configuration reported in the original Barlow Twins analysis: embedding dimention $d$ of 8192 with a $\lambda_{BT}$ value of 0.0051. The optimal linear probing accuracy was achieved with a $\lambda_{reg}$ value of 1.0.

\begin{table}[H]
\centering
\caption{Default hyperparameter configuration for Mixed Barlow Twins training on ImageNet with ResNet50 backbone.}
\label{tab:mbt-resnet18-in}
    \begin{tabular}{|c|c|}
    \hline
    \textbf{Key} & \textbf{Value} \\
    \hline
    \# gpus & 8 \\
    batch size & 1024 \\
    epochs & 300 \\
    learning rate biases & 0.0048 \\
    learning rate weights & 0.2 \\
    projector dims & ``8192-8192-8192''\\
    weight decay & 0.000001 \\
    $\lambda_{BT}$ & 0.0051 \\
    $\lambda_{reg}$ & 0.1 \\
    \hline
    \end{tabular}
\end{table}
{
\newpage
\bibliographystyle{ieeenat_fullname}
\bibliography{main}
}
\end{document}